\documentclass{article}
\usepackage{spconf,amsmath,epsfig}
\usepackage{subcaption, booktabs, amsmath, hyperref}

\usepackage{array}
\usepackage{placeins}

\usepackage{lineno,hyperref}
\title{URBAN CHANGE DETECTION USING A DUAL-TASK SIAMESE NETWORK AND SEMI-SUPERVISED LEARNING}
\name{Sebastian Hafner\textsuperscript{1,}\thanks{The research is funded by the Swedish National Space Agency, the project 'EO-AI4Urban' within the ESA and Chinese Ministry of Science and Technology's Dragon 5 Program, and Digital Futures within the project 'EO-AI4GlobalChange'.}, Yifang Ban\textsuperscript{1}, Andrea Nascetti\textsuperscript{1}}

\address{
	\textsuperscript{1 }Division of Geoinformatics, KTH Royal Institute of Technology, 114 28 Stockholm, Sweden\\
}
\begin{document}
%
\maketitle
\begin{abstract}

In this study, a Semi-Supervised Learning (SSL) method for improved urban change detection from bi-temporal image pairs is presented. The proposed method employs a Dual-Task Siamese Difference network that not only predicts changes with the difference decoder, but also segments buildings for both images with a semantic decoder. First, the architecture was modified to produce a second change prediction derived from the semantic predictions. Second, SSL was used to improve supervised change detection. For unlabeled data, we designed a loss that encourages the network to predict consistent changes across the two change outputs. The proposed method was tested on urban change detection using the SpaceNet7 dataset. SSL achieved improved results compared to three fully supervised benchmarks. Code for the paper is available at \url{https://github.com/SebastianHafner/SiameseSSL.git}.

\end{abstract}

\section{INTRODUCTION}

Urbanization is progressing at an unprecedented rate in many places around the world. Earth observation is a crucial tool to map land cover changes associated with urbanization. Change detection is typically conducted by comparing images acquired at different times that cover the same geographical area. Therefore, change detection is considered a binary classification problem \cite{ban2012multitemporal}.

In computer vision, the process of linking each pixel in an image to a class is referred to as semantic segmentation. Fully convolutional networks have become the state-of-the-art method for segmentation tasks by leveraging large collections of examples (i.e., labeled data). Fully convolutional networks such as the encoder-decoder architecture U-Net \cite{ronneberger2015u} can easily be adopted for change detection by concatenating bi-temporal image pairs along the channel axis, also referred to as Early Fusion (EF) \cite{daudt2018fully}. However, network architectures tailored to Earth observation change detection tasks may be more promising than off-the-shelf architectures adopted from computer vision. Daudt \textit{et al.} \cite{daudt2018fully} proposed two Siamese network architectures that pass images separately through two encoders with shared weights, before fusing the extracted information in a decoder. Several improvements to Siamese networks have since been proposed by, for example, learning more discriminative features from the input images \cite{liu2020building}.

Despite these efforts, current research on change detection with Siamese networks is predominately based on fully supervised learning; and thus, dependent on large collections of labeled data. However, change labels are scarce and manually annotating data is costly and time-consuming. On the other hand, unlabeled satellite data is plentiful. Therefore, it is desirable to investigate Semi-Supervised Learning (SSL) to improve supervised change detection by incorporate readily available unlabeled satellite data, alongside labeled satellite data, into network training \cite{zhu2009introduction}.

In this paper, we expand on a Siamese network architecture developed by others \cite{daudt2018fully,liu2020building} and propose a new learning task for unlabeled data. In combination with labeled data, we train the network using SSL and demonstrate its effectiveness by comparing it to several fully supervised benchmarks on the SpaceNet7 dataset \cite{van2021multi}.

\section{METHODOLOGY}

\subsection{The proposed method}
\label{subsec:proposed_method}

\begin{figure*}[ht]
    \centering
    \includegraphics[width=.95\textwidth]{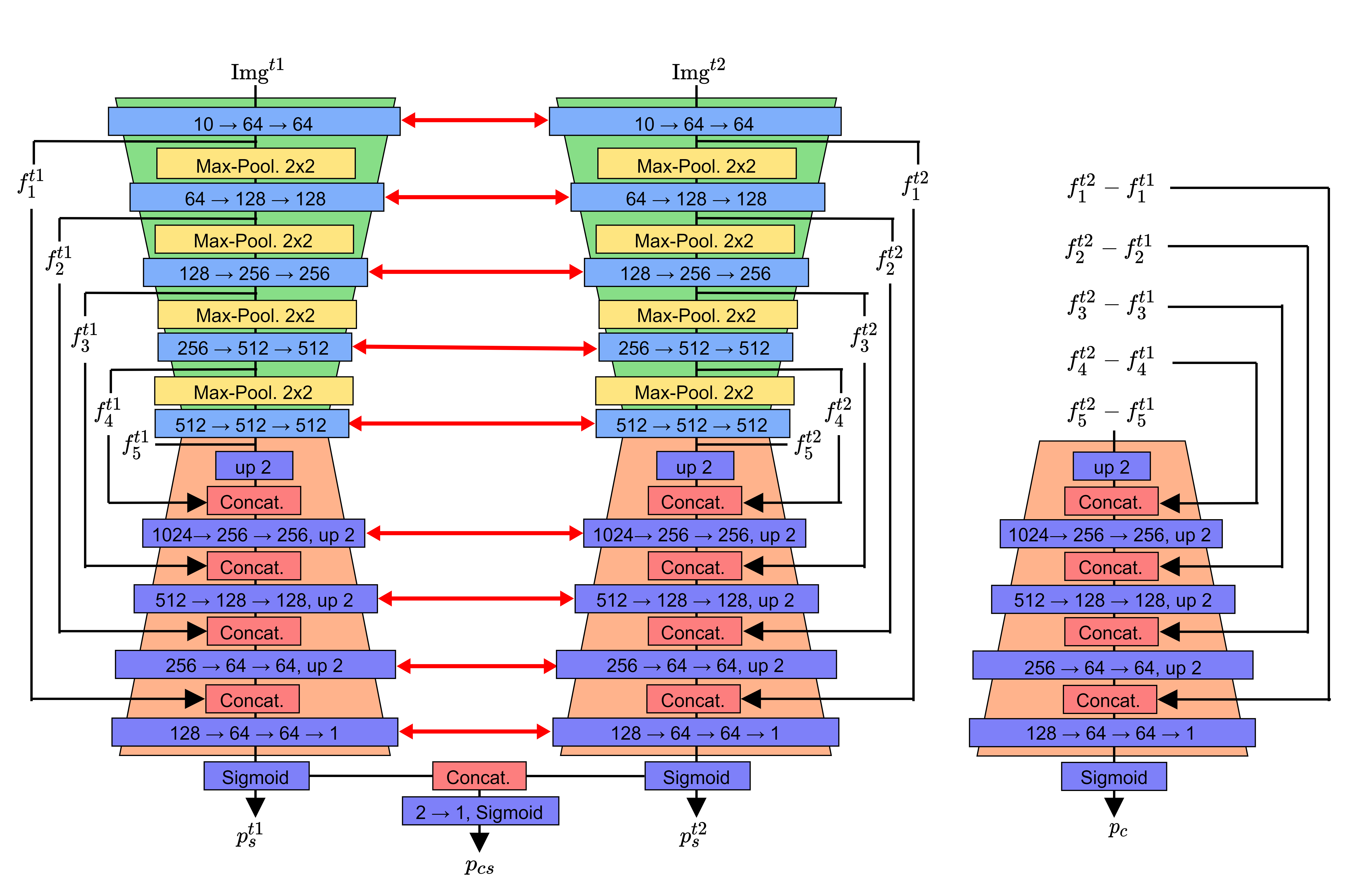}
    \caption{Diagram of the proposed Dual-Task Siam-Diff network for urban change detection with semi-supervised learning. Diagram style was adopted from \cite{daudt2018fully} where blue, yellow, red and purple blocks denote the operations convolution, max pooling, concatenation and transpose convolution, respectively. Red arrows illustrate shared weights.}
    \label{fig:architecture}
\end{figure*}

The proposed method consists of a new network architecture and a loss function to train the network in a semi-supervised fashion. The proposed network architecture (Fig. \ref{fig:architecture}) is an extension of the Siamese Difference (Siam-Diff) architecture introduced by Daudt \textit{et al.} \cite{daudt2018fully}. The Siam-Diff network processes images separately in encoders with shared weights (red arrows) to extract corresponding features ($f_1$--$f_5$) from image $t1$ and $t2$. The temporal features are then being forwarded via skip connections (black arrows) to the respective level of the difference decoder where they are subtracted from one another before being passed through subsequent layers. We incorporated the Dual-Task concept from Liu \textit{et al.} \cite{liu2020building} into our Siam-Diff network by adding decoders with shared weights for the semantic segmentation of buildings in images $t1$ and $t2$. Liu \textit{et al.} \cite{liu2020building} successfully used the Dual-Task concept to extract more discriminative features from input images with a Siamese network, which, in turn, improved change detection results. However, we use the building segmentation task not only to learn more discriminative features but also for SSL. To this end, the concatenated building outputs are passed through a 1x1 conv layer, generating an additional change output ($p_{cs}$). In total, our network produces four outputs: buildings segmentations for image $t1$ ($p_{s}^{t1}$) and $t2$ ($p_{s}^{t2}$), a change output from the building segmentation ($p_{cs}$) and the change output from the difference decoder ($p_{c}$).

The four network outputs are used to train the network in a semi-supervised fashion using a loss function composed of two supervised terms for labeled samples, namely for semantic ($\mathcal{L}_{s}$) and change ($\mathcal{L}_{c}$), and a consistency term ($\mathcal{L}_{cons}$) for unlabeled samples. The loss function is formally defined as follows:

\begin{equation}
\label{eq:loss}
\begin{aligned}
\mathcal{L}_{s} &= \mathcal{L}(p^{t1}_{s}, y^{t1}_{s})+\mathcal{L}(p^{t2}_{s}, y^{t2}_{s}) \\
\mathcal{L}_{c} &= \mathcal{L}(p_{c}, y_{c}) + \mathcal{L}(p_{cs}, y_{c})\\
\mathcal{L}_{cons} &= \mathcal{L}(p_{c}, p_{cs})\\
 \mathcal{L}_{sample} &=
    \begin{cases}
    \mathcal{L}_s + \mathcal{L}_c ,& \text{if } y \text{ exists} \\
    \varphi * \mathcal{L}_{cons}, & \text{otherwise}
\end{cases}
\end{aligned}
\end{equation}

where $s$ and $c$ denote semantic and change, respectively. Accordingly, $cs$ denotes change derived from the semantic outputs. For labeled samples, ground truth exists for the buildings at $t1$ ($y^{t1}_{s}$) and $t2$ ($y^{t2}_{s}$), and the derived change ($y_{c}$). The consistency loss term encourages change predictions to agree on unlabeled samples. Hyper-parameter $\varphi$ is used to tune its impact on the overall loss. Power Jaccard was used for all loss terms \cite{duque2021power}.




\subsection{Dataset}
\label{subsec:study_area}

\begin{figure}[h]
    \centering
    \includegraphics[width=0.48\textwidth]{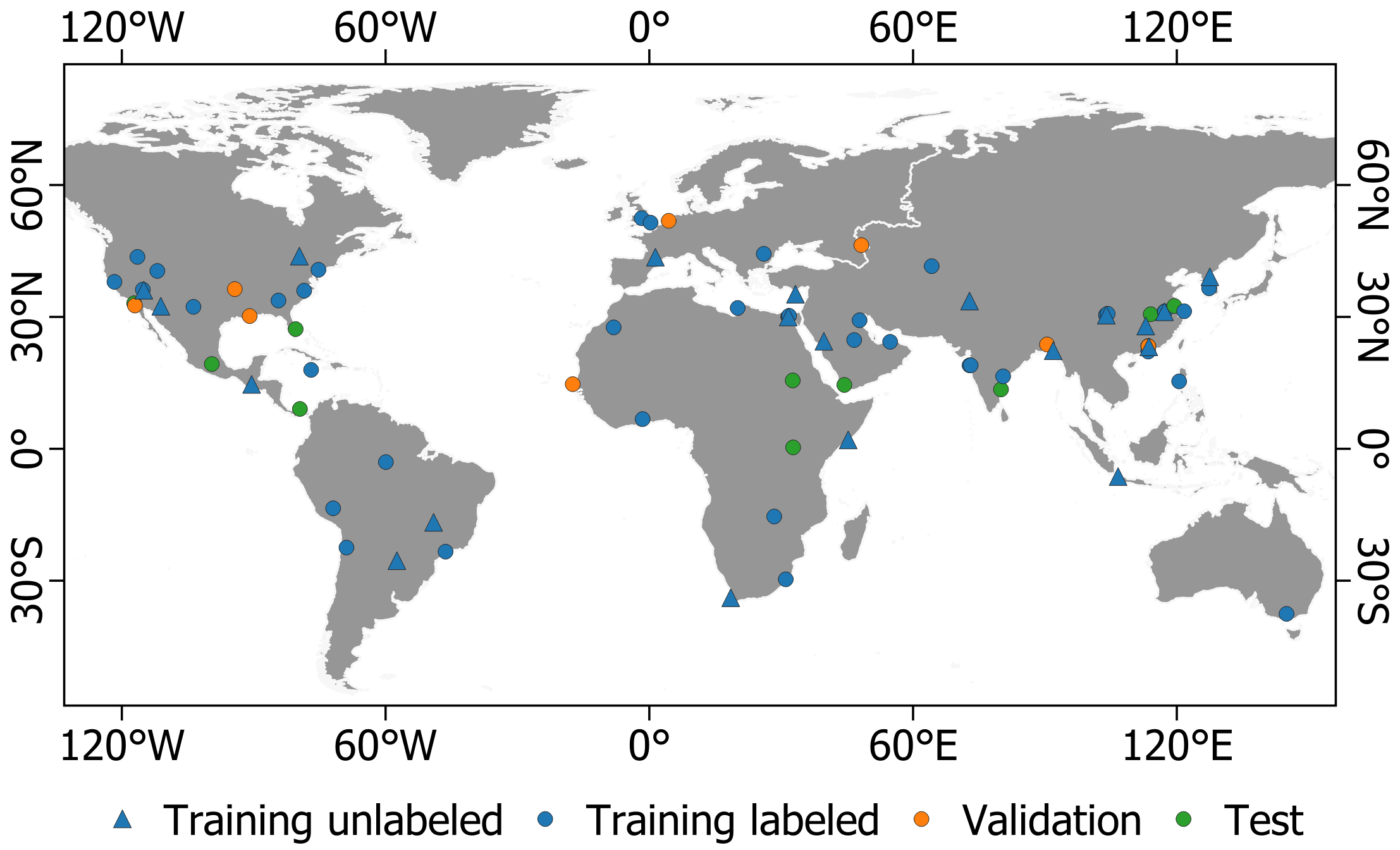}
    \caption{Overview of the SpaceNet7 sites constituting our training, validation and test set.}
    \label{fig:study_area}
\end{figure}

The proposed method was tested on the multi-temporal SpaceNet7 dataset \cite{van2021multi}. It consists of temporal stacks of monthly Planet composites, including corresponding manually annotated building footprints, ranging from the beginning of 2018 to 2020. An overview of the 60 SpaceNet7 train sites, split into training, validation and test set, and the  40 SpaceNet7 test sites used
as unlabelled data is shown in Fig. \ref{fig:study_area}. Timestamps affected by clouds were removed from the dataset.

\subsection{Network training and performance assessment}
\label{subsec:training_and_performance}

Training samples were generated by randomly selecting two timestamps from the time series of a site. The building labels for these timestamps, obtained from rasterizing the building footprints, were used to compute the change label. Twenty patches of size 256x256 pixels were then randomly cropped from the change label, before assigning each patch a probability according to its change pixel percentage, including a base probability for patches with no change pixels. A single patch was chosen based on those probabilities to oversample change pixels during training. Per epoch, one hundred samples were drawn from each site. The network was trained for 100 epochs with batch size 8 and an initial learning rate of 10\textsuperscript{-4}, using AdamW as optimizer. Horizontal and vertical flips and rotations ($k * 90^{\circ}$, where $k \in \{0, 1, 2, 3\}$) were used as data augmentations. Network training, implemented in PyTorch, was done on a NVIDIA GeForce RTX 3090 GPU. For the quantitative performance assessment, F1 score, precision (P) and recall (R) were used, defined as follows:

\begin{equation}
\label{eq:f1_score}
\begin{aligned}
    P = \frac{TP}{TP + FP} \:
    R = \frac{TP}{TP + FN} \:
    F1 = 2 * \frac{P * R}{P + R}
\end{aligned}
\end{equation}

where TP, FP and FN are true positives, false positives and false negatives, respectively.

\section{RESULTS AND DISCUSSION}

\begin{figure}[h]
    \centering
    \includegraphics[width=0.48\textwidth]{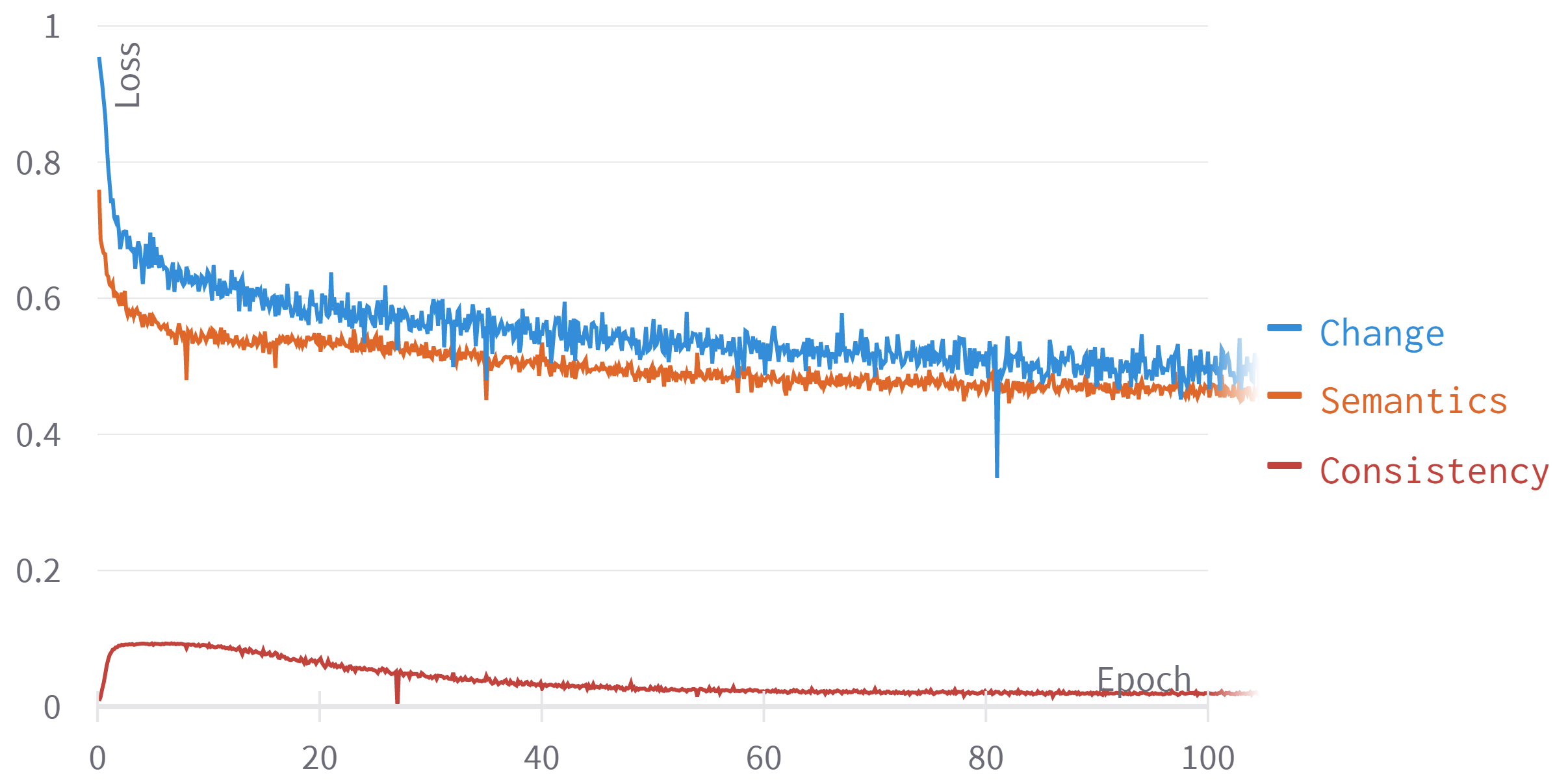}
    \caption{Loss terms comprising the loss function for the proposed change detection method using semi-supervised learning.}
    \label{fig:losses}
\end{figure}

Loss curves for the change, semantic and consistency term are shown in Fig. \ref{fig:losses}. At the beginning of training, losses for change and semantic rapidly decreased but then slowed down. In contrast, consistency loss first went up, before decreasing slowly. After approximately 50 epochs, all losses converged and remained relatively stable.

The performance of our proposed method was compared with three fully supervised benchmarks: EF U-Net \cite{daudt2018fully}, Siam-Diff \cite{daudt2018fully} and Siam-Diff + Dual-Task \cite{liu2020building}. To run a fair comparison, we adopted the concept from the papers but used encoder and decoder architectures identical to the ours (Fig. \ref{fig:architecture}), as well as an identical training setup (Section \ref{subsec:training_and_performance}). In terms of quantitative results (Table \ref{tab:quantitative_results}), our method achieved the highest F1 score on the test set (0.559), followed by Siam-Diff + Dual-Task (0.529). The lowest F1 score was obtained from Siam-Diff (0.484). For all networks, recall is good (0.650 +) but precision is relatively low with our method achieving the highest value (0.490). In contrast, the benchmarks did not exceed precision values of 0.450.

\begin{table}[h]
\small
  \caption{Quantitative change detection comparison of our method with benchmarks replicated using our encoder and decoder architectures and training setup.}
  \label{tab:quantitative_results}
  \centering
  \begin{tabular}{lrrr}
    \toprule
     & F1 score & Precision & Recall \\
    \midrule
    EF U-Net \cite{daudt2018fully} & 0.525 & 0.440 & 0.651 \\
    Siam-Diff \cite{daudt2018fully} & 0.484 & 0.368 & \textbf{0.704} \\
    Siam-Diff + Dual-Task \cite{liu2020building} & 0.529 & 0.440 & 0.664 \\
    Siam-Diff + Dual-Task + SSL & \textbf{0.559} & \textbf{0.490} & 0.651 \\
    \bottomrule
  \end{tabular}
\end{table}

\begin{figure*}[h]
    \centering
    \includegraphics[width=\textwidth]{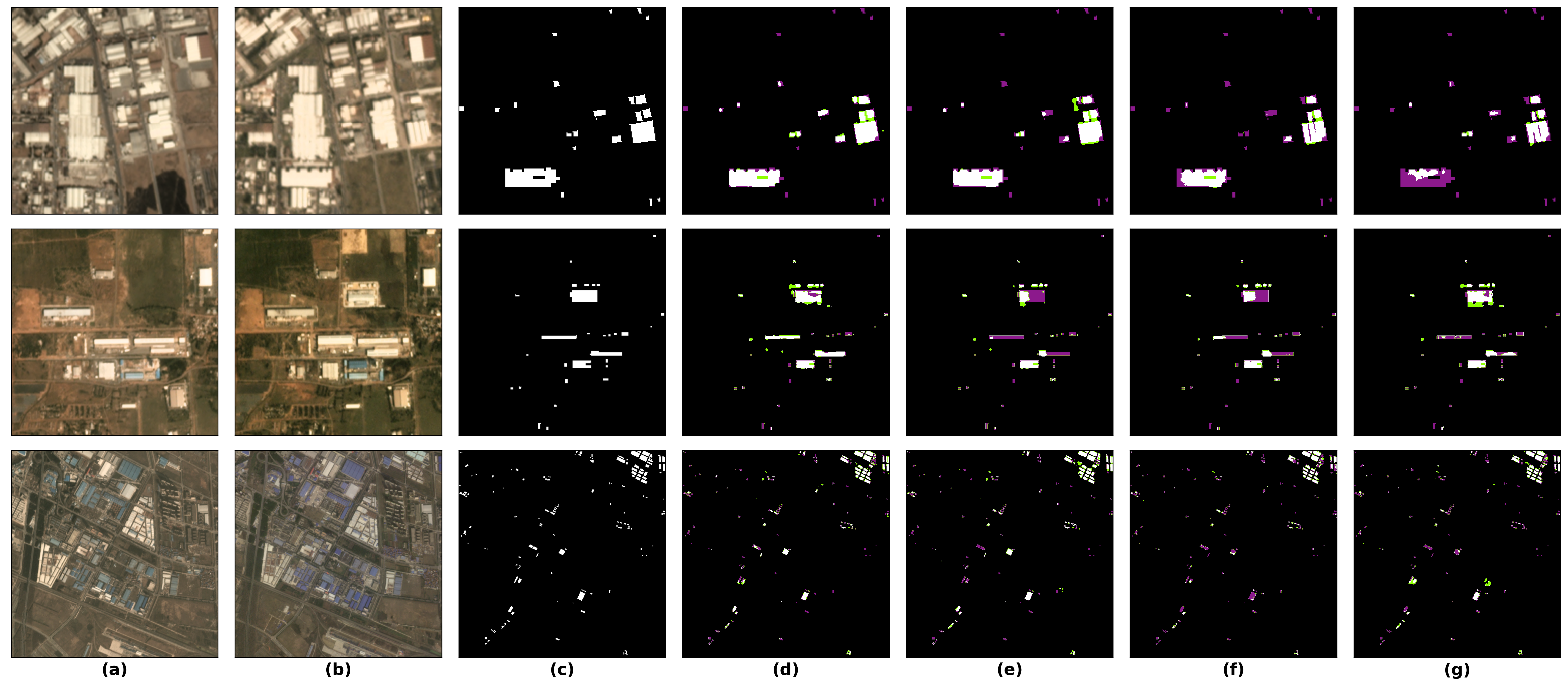}
    \caption{Qualitative comparison between the results obtained by (d) our Siam-Diff + Dual-Task + SSL method, (e) EF U-Net \cite{daudt2018fully}, (f) Siam-Diff \cite{daudt2018fully} and (g) Siam-Diff + Dual-Task \cite{liu2020building}. Images for $t1$ and $t2$ and the ground truth are shown in (a), (b) and (c), respectively. Legend for results: white is true positive, black is true negative, green is false positive and purple is false negative.}
    \label{fig:qualitative_comparison}
\end{figure*}

A qualitative comparison was done for sites in Mexico, India and China (Fig. \ref{fig:qualitative_comparison} from top to bottom). Generally, change detection ability of our network is better than that of the benchmark networks, indicated by fewer FNs (purple). However, a considerable amount of building constructions at small scale was not detected by any of the networks. Incorrectly detected changes (green) are mostly located around correctly detected change areas.

Despite improvements upon the supervised benchmarks, the proposed method is limited by the fact that both change outputs are based on the same underlying feature extraction. Therefore, encouraging consistency across change outputs fails to considerably improve the feature extraction; and thus, the learning potential from unlabeled data using the proposed consistency task may be limited. We assume that a more promising approach is to incorporate fusion of radar and optical data into SSL, since consistency across these data modalities is a powerful SSL task. Moreover, the different but complementary information in radar and optical data proved to be promising for urban change detection in combination with deep learning \cite{hafner2021sentinel}.

\section{CONCLUSION}

This study presents an improved method for urban change detection using a modified Dual-Task Siamese Difference network for SSL. In the proposed method, the two change outputs of our network are harnessed for a consistency learning task on unlabeled data. Experiments on the SpaceNet7 dataset showed that the proposed method improves upon fully supervised benchmarks. Future work will focus on incorporating multi-modal data into the proposed method.



\FloatBarrier
\bibliography{ref.bib}
\bibliographystyle{IEEEbib}

\end{document}